\documentclass[times,twocolumn,final]{article}

\usepackage{geometry}
\usepackage{framed,multirow}

\usepackage{amssymb}
\usepackage{latexsym}
\usepackage{graphicx}
\usepackage[style=numeric,backend=biber,bibencoding=utf-8]{biblatex}
\usepackage{float}

\usepackage{url}
\usepackage{xcolor}
\definecolor{newcolor}{rgb}{.8,.349,.1}

\usepackage{hyperref}
\usepackage{tabularx}
\usepackage{subcaption} 
\usepackage[switch,pagewise]{lineno}

\usepackage[frozencache]{minted}
\setminted{fontsize=\small,baselinestretch=1}
\usepackage{mathtools}

\usepackage{subcaption} 
\usepackage{tablefootnote}
\usepackage[switch,pagewise]{lineno}

\usepackage{mathtools}
\usepackage{array}

\newcommand\extrafootertext[1]{%
    \bgroup
    \renewcommand\thefootnote{\fnsymbol{footnote}}%
    \renewcommand\thempfootnote{\fnsymbol{mpfootnote}}%
    \footnotetext[0]{#1}%
    \egroup
}

\begin{document}

\title{An Indexing Scheme and Descriptor for 3D Object Retrieval Based on Local Shape Querying}

\author{Bart Iver van Blokland \and Theoharis Theoharis}

\maketitle

\begin{abstract}
    A binary descriptor indexing scheme based on Hamming distance called the Hamming tree for local shape queries is presented. A new binary clutter resistant descriptor named Quick Intersection Count Change Image (QUICCI) is also introduced. This local shape descriptor is extremely small and fast to compare. Additionally, a novel distance function called Weighted Hamming applicable to QUICCI images is proposed for retrieval applications. The effectiveness of the indexing scheme and QUICCI is demonstrated on 828 million QUICCI images derived from the SHREC2017 dataset, while the clutter resistance of QUICCI is shown using the clutterbox experiment. 
\end{abstract}  

\extrafootertext{\copyright2020. This manuscript version is made available underthe CC-BY-NC-ND 4.0 license \url{http://creativecommons.org/licenses/by-nc-nd/4.0/}}

\section{Introduction}
\label{sec:introduction}

The problem of shape retrieval has thus far primarily been posed as an object based one. Many proposed algorithms aim to answer queries such as `find all chairs', or `find buses similar to this sample bus'. However, objects are not a single large shape; they are the sum of many small details that combined produce a larger, more complex whole. For instance, a wheelbarrow may contain shapes such as a slightly bent flat surface, a curved cylinder, or a large disc. Individually these shapes may not be unique to that object, but their specific combination and arrangement makes it an object useful for garden work. 

It may be argued that querying of such smaller (partial) shapes fall under the existing class of partial object retrieval. Thus one can pose retrieval queries such as 'find all objects that contain a spout like this', which would presumably retrieve teapots (as well as other objects with spouts). Unfortunately, partial retrieval requires the availability of the partial shape that is to be retrieved. 

However, in many cases it is useful to be able to pose more general geometric queries such as 'retrieve objects containing an S-bend' for finding bottles with that specific cross-section. This could be easily specified by drawing such a curve in 2D thus not necessitating the existence of a partial query object.

One important problem with this type of approach, which would have to describe shape at a very low level, is the sheer volume of local descriptors that would be generated, potentially one for every vertex. Not only would they require a large amount of storage but it would also be quite slow to search them. 

We thus propose:

\begin{itemize}
    \item A robust and efficient novel local binary shape descriptor (called QUICCI)
    \item An efficient novel indexing scheme called Hamming Tree for bit strings such as QUICCI
    \item A novel distance function used for retrieval of bit strings (called Weighted Hamming distance)
\end{itemize}

After an introduction to relevant background material in Section \ref{sec:background}, each of these contributions are described in the above listed order in Sections \ref{sec:quicci}, \ref{sec:hammingTree}, and \ref{sec:weighted_hamming}, respectively. The various methods are evaluated in Section \ref{sec:evaluation}, and some specifics are discussed in Section \ref{sec:discussion}.

\section{Background and Related Work}
\label{sec:background}

This section is divided in two parts, corresponding to the main contributions of the paper: indexing bit strings and local shape descriptors.

\subsection{Indexing Bit Strings}

The need for indexing collections of bit strings primarily stems from two main categories of methods; those utilising dimensionality reduction to map higher dimensional descriptors on to shorter binary vectors, and binary feature descriptors. 

Dimensionality reduction is often done through a method utilising Locality-Sensitive Hashing (LSH), initially described by Har-Peled et al. \cite{Har-Peled_Indyk_Motwani_2012}. These aim to represent larger, more varied feature vectors in shorter bit strings, where similar feature vectors will produce similar bit strings, thereby significantly reducing the search space for finding closest neighbours. Popular methods applying LSH include Minhash proposed by Broder et al. \cite{broder1997on} (as well as a more scalable variant \cite{broder2000identifying}), and Simhash \cite{sadowski2007simhash} by Sadowski et al.

A number of binary feature descriptors have been proposed aimed at various retrieval applications. For images, the most popular binary features proposed to date include BRIEF by Calonder et al. \cite{calonder2010brief}, a rotation invariant extension named  ORB by Rublee et al. \cite{rublee2011orb}, and a both rotation and scale invariant keypoint descriptor called BRISK by Leutenegger et al. \cite{leutenegger2011brisk}. An example of a binary descriptor for 3D point matching is B-SHOT, proposed by Prakhya et al. \cite{prakhya2015b}. The lengths of these descriptors vary between 128 and 512 bits.

While LSH derived methods are capable of significantly reducing dimensionality in the source data, large quantities of indexed data may cause a high number of hash bins to be created. However, not all hash bins may receive a similar number of entries, and the creation of all possible bins for a given bit string length is not always feasible, thereby creating the need to discover the existence of nearby hash bins with relatively low Hamming distances. This discovery process can be costly, particularly when no close neighbours exist. Binary feature descriptors are inherently longer, and for that reason face a similar problem.

Retrieval from large collections of bit strings, where each bit string is ranked by hamming distance from a query string, is known in the literature as the \emph{n Nearest Neighbours} Hamming problem, and a variety of methods have been proposed \cite{brodal1996approximate} \cite{brin1995near} \cite{arslan2004dictionary} \cite{maai2007text}.

However, these early methods are limited in their design to the efficient retrieval of neighbours with Hamming distances of up to 2, support for short bit sequences only, or both. More recent methods have addressed the problem more effectively, and do not exhibit the aforementioned problems.

Norouzi et al. proposed the Multi-Index Hashing (MIH) algorithm \cite{norouzi2012fast}. The method works by dividing all indexed bit strings into equally sized disjoint substrings, and constructing a hash table for each set of corresponding substrings. These can be queried by subdividing the query string in a similar fashion, and querying each hash table for all permutations of the query substring within a given Hamming distance. The set of candidate matches can be refined when testing subsequent hash tables, as strings which surpass the Hamming distance limit can be excluded prematurely. The authors show that MIH outperforms the most significant previous work, however, the requirement that all permutations within a given Hamming distance must be tried on hash sets becomes a performance bottleneck when this limit is high.

Chappell et al. proposed a system for approximate nearest-neighbour search of bit strings \cite{chappell2015approximate} aimed at locating such nearest neighbour hashes by creating inverted lists of smaller bit string ``slices'', similar to the divisions done in MIH. However, for larger collections of longer bit strings, such as binary descriptors, the method does not scale due to each slice list increasing in size linearly.

Reina et al. \cite{reina2017an} presented an improved variation of MIH. This is a hybrid indexing scheme, where a \emph{trie} (prefix tree) is used to store the index tree itself, and a separate hash table is exploited to prune tree branches during a query by checking a specific bit string's existence in the index when the tree node's common prefix has reached a given Hamming distance limit. In similar fashion to MIH, bit strings are divided into substrings, and from each corresponding substring a separate index is constructed.
While the method is shown to outperform MIH, it is hampered by the fact that for its efficiency (the pruning of branches which are known not to contain matches) it relies on the existence of a fixed Hamming distance limit. Constructing a querying algorithm which does not contain this optimisation is possible, but as the authors themselves state, this would significantly degrade querying performance. Moreover, creating one index for each subdivision in the input string, as well as the corresponding hash table and trie that each of these includes, adds significant storage overhead. 

The Hamming Tree data structure proposed in this paper commands a significantly lower overhead, as only a single indexing structure is created. The proposed querying algorithm can dynamically cut off the querying process and does not necessarily require a Hamming distance limit to be set.

\subsection{Local 3D Shape Descriptors}

Local approaches to 3D object retrieval are advantageous to global methods due to their inherent resistance to clutter and shape variations. The field is well developed, and numerous descriptors have been proposed to date, e.g. \cite{tombari2010unique} \cite{flint2007thrift} \cite{guo2013rotational} \cite{giachetti2012radial}. 

One popular descriptor is the Fast Point Feature Histogram (FPFH) \cite{rusu2009fast}. It is constructed in two phases. First, a Simplified Point Feature Histogram (SPFH) is computed for each point in the scene, by constructing a Darboux frame to each neighbour in the point's vicinity, and accumulate its components over a fixed number of bins. Next, the FPFH descriptor of a point is constructed by adding the average SPFH histogram of the point's neighbours (albeit also weighted by distance to the point itself) to the point's own SPFH.

While many such descriptors have been shown to perform well at recognition tasks, one of their primary challenges is the presence of geometry unrelated to the queried shapes within the support volume of a descriptor, referred to as ``clutter'' \cite{guo2016comprehensive}. Not every descriptor is equally resistant to the negative effects of clutter on matching performance. 

One example of a descriptor that has been shown to resist clutter is the classic Spin Image (SI) proposed by Johnson et al. \cite{johnson1999using}. An SI is generated by projecting points uniformly sampled from a surface on to a rotating square plane whose side is on the axis of rotation. The plane is divided into square bins, which count the number of point samples projecting onto them, thus creating a 2D histogram. Similar surfaces will result in a high correlation between their Spin Images. 

An extension to the Spin Image which is related to the descriptor proposed in this paper, is the Spin Contour descriptor proposed by Liang et al \cite{liang2015geodesic}. The authors post-process high resolution Spin Images by detecting edges between zero and nonzero histogram bins. The resulting outlines, called ``Spin Contours'', are used for shape detection. However, the Spin Contour can only be used to represent an object in its entirety, due to its inability to detect edges within a Spin Image. Moreover, due to the method's reliance on outlines, its clutter resistance is not expected to be good.

Another descriptor shown to be resistant to clutter is the 3D Shape Context (3DSC) proposed by Lowe et al \cite{lowe2004distinctive}. 3DSC has a spherical support volume, which is subdivided into bins through horizontal and vertical cuts, as well as spheres placed within it. Point samples of the surrounding region intersecting each bin are subsequently accumulated, creating a histogram. 3DSC descriptors are compared using a Euclidean distance function.

The Radial Intersection Count Image (RICI) \cite{vanrici} is a descriptor aimed at shape matching in highly cluttered scenes. A set of three dimensional circles are defined with centers along the normal to a vertex and with varying radii. The number of intersections between each circle and the mesh surface is counted, resulting in a 2D histogram. This is similar to the arrangement of circles seen in Figure \ref{fig:quicci_construction}. Comparing exact changes in intersection counts from a circle to its neighbour can be used to determine correspondence between RICI descriptors. The authors also propose a distance function that is capable of largely disregarding clutter within the support region, and show that this results in better matching performance in heavily cluttered scenes.

\section{Quick Intersection Count Change Image (QUICCI)}
\label{sec:quicci}

In contrast to the previously proposed RICI descriptor, which stores integers representing intersection counts, the QUICCI descriptor stores booleans representing changes in intersection counts. The differences between the two descriptors also propagate to their distance functions, which due to the different representations require them to be tailored specifically to each method.

\begin{figure}
    \centering
    \includegraphics[width=8.8cm]{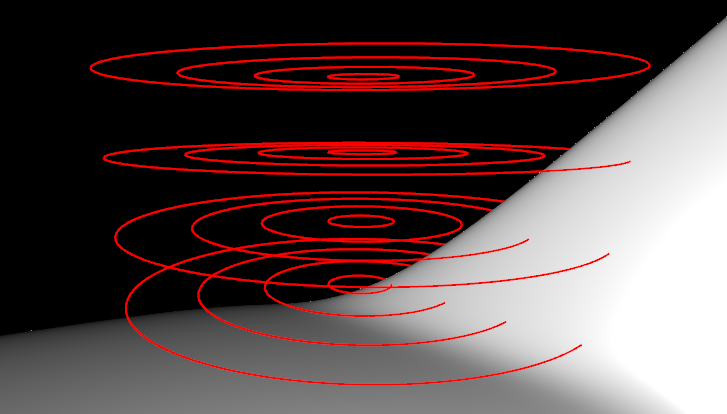}
    \caption{\small Visualisation of the ``layers of circles'' used for the construction of a 4x4 RICI, or a 3x4 QUICCI descriptor (pairs of circles are used, thus the effective width is one less than the number of circles per layer). The 4 layers containing 4 circles each can be seen in the image, some of which intersect with the object surface towards the right side. The circles combined form a cylindrical support volume.}
    \label{fig:quicci_construction}
\end{figure}

Circles are arranged in layers, each layer containing circles of increasing radii by a constant amount increment. The distance between circle layers, and the radius increment between circles within a layer, are equal. One thus forms a cylindrical ``grid'', where the y-coordinate corresponds to a layer of circles, and the x-coordinate to a circle within that layer. These coordinates in turn can be used to create an image. A visualisation of this is shown in Figure \ref{fig:quicci_construction}.

The descriptor is constructed around an oriented point, consisting of a vertex and a normal, referred to as the Reference Vertex and Reference Normal for the remainder of this paper. The oriented point defines a three-dimensional line, called the Central Axis. All circles are orthogonal to the Reference Normal, and centred around the Central Axis. The Reference Vertex lies at the exact centre of the support region.

\begin{figure}
    \centering
    \includegraphics[width=8.8cm]{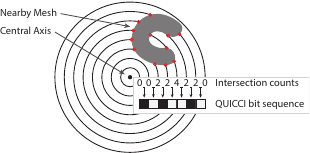}
    \caption{\small Visualisation of the construction of a single row of a QUICCI descriptor. First, intersections between circles with increasing radii and a mesh surface are counted (intersection points are indicated in red). Next, neighbouring intersection counts are compared. If they are different, the corresponding bit in the QUICCI image is set to 1 (white), otherwise to 0 (black).}
    \label{fig:quicci_layer_construction}
\end{figure}

Computing a QUICCI descriptor for a Reference Vertex involves intersecting all circles with the mesh surface, and subsequently subtracting each circle's intersection count from that of its smaller neighbour in the same layer, as illustrated in Figure \ref{fig:quicci_layer_construction}. If this difference is nonzero, the corresponding bit will be set to 1, else to 0. This implies that a layer with $C$ circles will result in a QUICCI descriptor of $C - 1$ bits.

The function for comparing two QUICCI descriptors is asymmetric, and distinguishes between a needle image (describing the shape that should be located) and a haystack image (describing any other shape to which a similarity score should be computed). Intersection count changes present in the needle image are characteristic to the shapes being queried, and this can be exploited by only including those specific bits in the distance computation. Due to the QUICCI image's tendency to be sparse, this excludes the majority of the image's bits from the distance computation, making it resistant to clutter (see Section \ref{sec:quicciPerformance}). An algebraic representation of the distance function is shown in Equation \ref{eq:quicciDistanceFunction}.

\begin{equation}
\label{eq:quicciDistanceFunction}
    D_{QUICCI}(I_n, I_h) = \sum_{r = 0}^{N} \sum_{c = 0}^{N} ((I_n[r,c] \oplus I_h[r,c]) \wedge I_n[r,c])
\end{equation}

Where $I_n$ and $I_h$ are the needle and haystack images, respectively, $I[r,c]$ denotes the bit at row $r$ and column $c$ of image $I$, and $N$ denotes the QUICCI image width. A lower distance value indicates that the two images are more similar. The $\oplus$ and $\wedge$ operators denote the bitwise XOR and AND functions, respectively.

Constructing the QUICCI descriptor as a binary image has significant advantages. Many of the previously discussed local shape descriptors use 32-bit floating point or integer representations. The QUICCI descriptor thus uses about an order of magnitude less memory.  This smaller size means both less disk storage and significantly faster comparison rates, mainly due to the smaller memory bandwidth requirements.

Moreover, QUICCI descriptors can be constructed efficiently on the GPU due to its advantageous memory access patterns and bandwidth requirements. The intersection computation between the circles and the mesh surface is the most demanding part of this process, which can be done using the efficient algorithm presented in \cite{vanrici}.

\section{Hamming Tree}
\label{sec:hammingTree}

The Hamming Tree is introduced as a means for indexing bit strings of arbitrary length, such as QUICCI images, for the purpose of k-Nearest-Neighbour searches using the Hamming distance function \cite{Hamming_1950} as a ranking metric. In this paper, the method is discussed and tested only on the proposed QUICCI descriptor, where the rows of the complete 2D image are concatenated to produce a 1D bit string that can be indexed and queried. However, the application of the tree is not limited to it and can be used for indexing arbitrary bit strings. For this reason the explanations in this section will use QUICCI images as an example, but the contents of the tree being indexed is referred to as ``bit string'' throughout the paper.

The observation central to the design of the tree is that the total set bit count of a bit string can be used to compute a lower bound of the Hamming distance between a given pair of bit strings. For example, two bit strings with 3 and 5 bits set, respectively, must have a Hamming distance of at least 2. This minimum distance can subsequently be used as a heuristic for navigating a tree, where the set bit count of consecutive, equally sized, substrings determines which next branch to pick. Each branch taken will place stricter requirements on the distribution of set bits within the string, increasing the probability a match is found.

\subsection{Hamming Tree Construction}

Construction of the tree is done by iteratively inserting bit strings, dynamically expanding the tree where necessary. It consists of two node types; internal nodes and leaf nodes. Internal nodes contain references to leaf nodes and other internal nodes. Leaf nodes in turn contain a list of bit strings. Initially, the tree consists of one root internal node, and one leaf node for each possible bit count. When the length of all bit strings being indexed is $N$, that implies the root node has $N + 1$ children. For levels underneath the root node, the branching factor is at most the number of set bits corresponding to that node plus one.

\begin{figure}
    \centering
    \includegraphics[width=8.8cm]{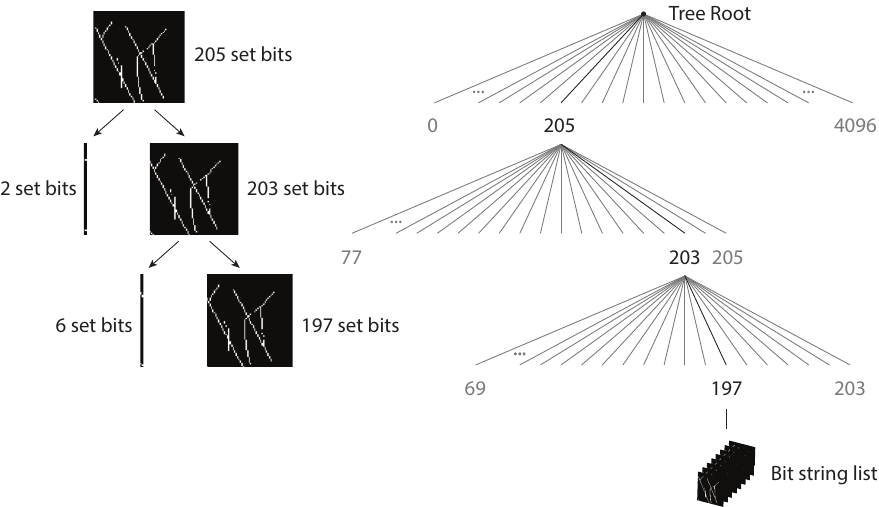}
    \caption{\small Visual representation of navigating a Hamming tree. On the left hand side, a $64 \times 64$ QUICCI image is shown, where two columns are removed at each step (128 bits). The right hand side shows the corresponding path in the tree.}
    \label{fig:tree}
\end{figure}

The tree is navigated (both during insertion and querying) by iteratively cutting off a fixed number of bits from the front of the bit string. After removal, the number of set bits in the remaining string determines the next branch to take in the tree. This process has been visualised in Figure \ref{fig:tree}. While this approach does not place requirements on the exact positions of set bits, it aims to filter the indexed bit strings by those whose distribution of bits are similar to a given query string, thereby increasing the likelihood a relevant match is found. 

Thus to insert a new bit string into a Hamming tree, one navigates down to the leaf node corresponding to the bit string. The new bit string is then inserted in the list of that leaf node. If afterwards the count of that list exceeds a constant threshold, the leaf node is replaced by an internal node and the list of bit strings is distributed among the lists of its new child leaf nodes.

\subsection{Querying the Hamming Tree}

Our algorithm for querying a Hamming tree takes in a needle bit string, a Hamming tree, and the maximum number of search results that should be returned as input, and returns a list of bit strings whose Hamming distance is closest to the provided needle string.

It first attempts to locate an exact match for the needle string and subsequently widens the search so as to include the nearest matches within the requested search result limit count. 

Internally, the algorithm maintains a list and a priority queue. The list stores the best search results that have been found up to that point and its size is limited by the search result limit parameter. The priority queue contains unvisited internal tree nodes (initialised with the tree's root node), sorted by the minimum possible distance between the needle and all possible descriptors in the subtree rooted at a node.

\begin{figure}
    \centering
    \includegraphics[width=8.8cm]{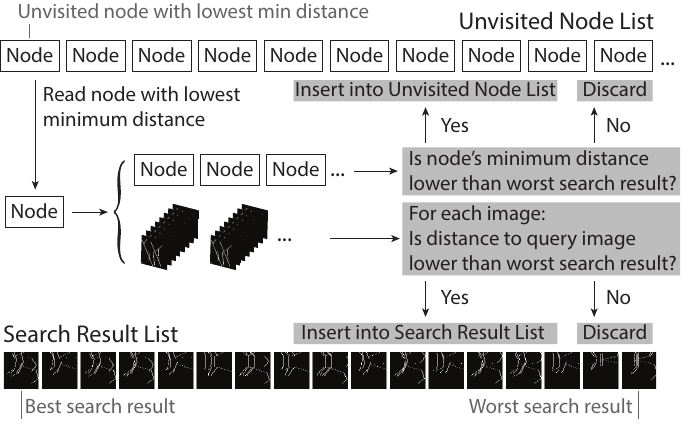}
    \caption{\small Visualisation of the Hamming tree query algorithm. At each iteration, the contents of the node with the lowest minimum distance in the unvisited node queue, which consists of child internal nodes and bit string lists, is inserted into the univisited node list and search result list, iff there is a possibility that they can potentially improve the search results.}
    \label{fig:query_algorithm}
\end{figure}

The query algorithm visits one internal node at a time, until the node at the front of the unvisited node queue (with the lowest minimum distance) has a greater distance than the worst entry in the search result list, or the unvisited node queue is empty (a similar strategy to the one adopted by Chappell et al. \cite{Chappell_Geva_Nguyen_Zuccon_2013} \cite{chappell2015approximate}). This process is illustrated in Figure \ref{fig:query_algorithm}.

Visiting an internal node involves iterating over the node's outgoing edges. When there is a bit string list at the end of that edge, compute the Hamming distance between the needle and haystack strings contained within. Any strings which improve the list of search results are inserted into the search result list. When the outgoing edge points to an internal node, the minimum distance to that node is computed (as a function of the needle string and the node's position in the tree), and if that minimum distance is lower than the current worst entry in the search result list, it is inserted in the unvisited node queue. This condition prevents the unvisited node queue from growing indefinitely, and excludes bit strings that are certainly not going to be part of the search results anyway.

The Hamming Tree is capable of efficiently locating all bit strings which have low distance scores relative to a needle string. However, as the distances get larger than a few bit flips, the number of permutations, and thus nodes that need to be visited, increases to such a degree that it may be necessary to visit a significant part of the tree before the algorithm can ensure that no better search results exist. It is therefore advisable to set a distance threshold that is used in conjunction with the worst search result score, and set this threshold as low as possible when querying a Hamming tree.

In terms of complexity, in the worst case, a completely unbalanced tree is in effect equivalent to a linked list. Insertion therefore has constant complexity ($O(1)$), while search is linear ($O(n)$).

\section{Weighted Hamming Distance}
\label{sec:weighted_hamming}

\begin{figure}
    \centering
    \includegraphics[width=6cm]{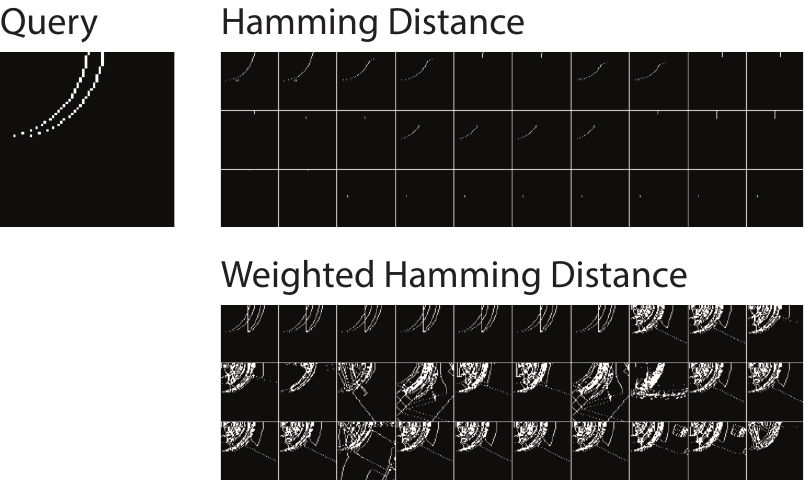}
    \caption{\small Top 30 search results for the shown needle image when search results are ranked using Hamming distance (above) and the proposed Weighted Hamming (below) distance functions.}
    \label{fig:distance_function_comparison}
\end{figure}

With respect to QUICCI, for a proportion of needle images, the previously proposed indexing strategy is capable of locating QUICCI images containing similar shapes as to the ones requested in the needle. However, it is not universally applicable for this task. Most notably, needle images which are close to fully saturated with set (1) or unset (0) bits are likely to yield search results containing irrelevant shapes. An example of such a needle image and the corresponding search results can be seen in Figure \ref{fig:distance_function_comparison}. The needle image shown in the Figure is also a good example of a local shape query of the kind described in Section \ref{sec:introduction}. 

The cause of this problem is that the Hamming distance considers each bit to be equivalent in importance. However, when the number of set bits in a needle image is low, for the purpose of shape retrieval, it is more important that the bits set in the needle are also set in the haystack image than unset needle bits being unset in the haystack image. We therefore observe that the lower the number of set bits in the needle image, the more important it is for these bits to be set in a haystack image. The opposite also holds true for needle images nearly saturated with set bits.

With this observation, we propose an alternate distance function, called ``Weighted Hamming'', which can be used for ranking QUICCI search results. The function broadly resembles Hamming distance, but the distance cost for the two types of bit mismatches (incorrectly set and incorrectly unset) are weighted differently, depending on the proportion of set to unset bits in the needle image. The definition of this function is given in Equation \ref{eq:distanceFunction}.

\begin{align}
\begin{split}
    \label{eq:distanceFunction}
    D_{WH}(I_n, I_h) &=  \frac {\sum_{r = 0}^{N} \sum_{c = 0}^{N} (I_{n}[r,c](1 - I_h[r,c]))} {max(\sum_{r = 0}^{N} \sum_{c = 0}^{N} I_{n}[r,c], 1)} \\
    &+ \frac {\sum_{r = 0}^{N} \sum_{c = 0}^{N} ((1 - I_{n}[r,c])I_h[r,c])} {max(N - \sum_{r = 0}^{N} \sum_{c = 0}^{N} I_{n}[r,c], 1)}
\end{split}
\end{align}

Here $I_n$ and $I_h$ are respectively the needle and haystack images being compared, $I[r,c]$ represents the bit at row $r$ and column $c$ of a needle or haystack image $I$, and $N$ is the width of the QUICCI image in bits. 

It's worth noting that removing the denominators of the fractions in the Equation makes it equivalent to the regular Hamming distance function. Moreover, the weighted Hamming distance function is effectively a hybrid between Hamming distance and the clutter-resistant QUICCI distance function used for locating shapes in clutter heavy scenes shown in Equation \ref{eq:quicciDistanceFunction}. When the second term in Equation \ref{eq:distanceFunction} is nullified, its ranking becomes equivalent to the clutter-resistant distance function. However, the removal of the second term also means there is a possibility for false positives, where a high variation in intersection counts may cause the desired needle bits to be set accidentally in a given haystack image, even though the surroundings of the corresponding haystack point does not actually contain the shapes requested by the query. We therefore consider the function given in Equation \ref{eq:distanceFunction} to be more suitable for retrieval purposes. An in-depth evaluation of this claim is done in Section \ref{sec:whammingEvaluation}.

\subsection{Indexing for Weighted Hamming}

The remainder of this section is dedicated to the construction of an index that allows querying using the presented weighted Hamming distance function, and a discussion of insights and some negative results that were obtained in the process. It is assumed that needle images will generally have a low number of bits set, otherwise a regular Hamming tree is likely a more suitable solution.

A good indexing strategy is closely tied to the distance function used. For weighted Hamming, this means that since the function primarily looks for the bits which are set in the needle image, the indexing structure should focus on discovering those in haystack images. One observation that can be made for QUICCI images is that edges of 3D geometry tend to create line or curve responses in QUICCIs. Thus, groups of bits that are in close proximity to one another in a QUICCI image are likely to be related. 

There are a number of ways in which this can be exploited, however, a problem is the exponential increase of permutations in the possible arrangements of a group of bits. However, it can be observed that due to the image's construction, these lines have a tendency to be vertical. This allows a relatively simplistic approach where permutation counts remain within reasonable limits.

The indexing algorithm detects segments of consecutively set bits within each column of the QUICCI image. Vertical bit sequences are advantageous due to their aforementioned common occurrence, and limited number of permutations in which they can occur within a given column. 

\begin{figure}
    \centering
    \includegraphics[width=8.8cm]{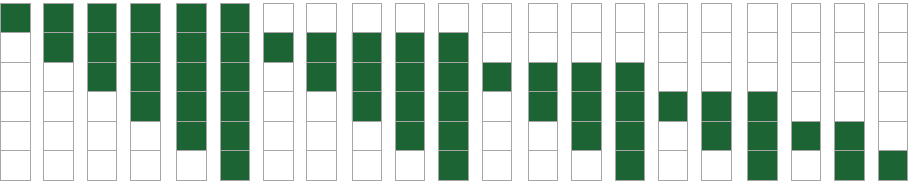}
    \caption{\small All 21 possible sequences of consecutively set bits in a single column of a 6 bit high image. White represents unset bits, whereas those marked green are set. A single column may contain multiple (albeit non-overlapping and separated by at least one unset bit) such sequences.}
    \label{fig:pattern_columns}
\end{figure}

For every possible bit sequence, an inverted list is created of all images which contain that exact bit sequence (with the same starting position and length). As an example, all possible consecutive bit sequences that can be found in an image that is 6 bits high are shown in Figure \ref{fig:pattern_columns}.

Querying involves combining the contents of all lists whose bit sequence overlaps by at least 1 bit with the given needle image. Since the total number of set bits for each image is also stored with each entry in the inverted lists, the exact weighted Hamming distance can be computed and used to rank results.

Unfortunately, the major issue of this approach is also the main advantage of the Weighted Hamming function; the value of matching set bits between the needle and haystack images far outweighs the cost of mismatched unset needle bits. The query algorithm must therefore consider all haystack bit sequences that overlap by at least one bit with the needle image, and cannot preemptively disregard entries. This causes many inverted lists to be searched, incurring long query execution times (typically resulting in a cost similar to a sequential search). Furthermore, the storage requirements of this index are high due to the inverted list construction.

\section{Evaluation}
\label{sec:evaluation}

All experiments involving time measurements in this section were executed on the same hardware. For CPU implementations, an Intel Xeon Platinum 8168 was used, and GPU kernels were executed on an NVidia Tesla V100 SXM3. The remainder of the results were in part gathered on the NTNU IDUN Cluster \cite{sjalander+:2019epic} computing cluster. 

\subsection{Hamming Tree Search Acceleration}
\label{subsec:hammingTreeResults}

The Hamming Tree was implemented in C++. Nodes and image lists stored on disk are compressed using the LZMA2 algorithm \cite{fastlzma2}. This was selected after empirically testing a number of state-of-the-art compression methods; LZMA2 yielded good compression ratios and speed for QUICCI images.

A Hamming Tree was constructed over the first 12,500 objects of the SHREC2017 dataset \cite{savva2017shrec}, which resulted in a total of 828.5 million QUICCI images. The resolution of the QUICCI images was set to 64x64 bits, and the support radius to 0.3 (for consistency all objects are translated and scaled to fit into a unit sphere prior to QUICCI generation). The number of bits removed at each level of the Hamming tree was set to 128 bits, or 2 image columns. The threshold at which leaf nodes (image lists) are split was set to 256 images, which balances index compactness with granularity. 1000 queries were executed on the constructed Hamming Tree. The needle images were randomly selected from the entire SHREC2017 QUICCI dataset (51,109 objects). 

\begin{figure}[h]
    \centering
    \includegraphics[width=8.8cm]{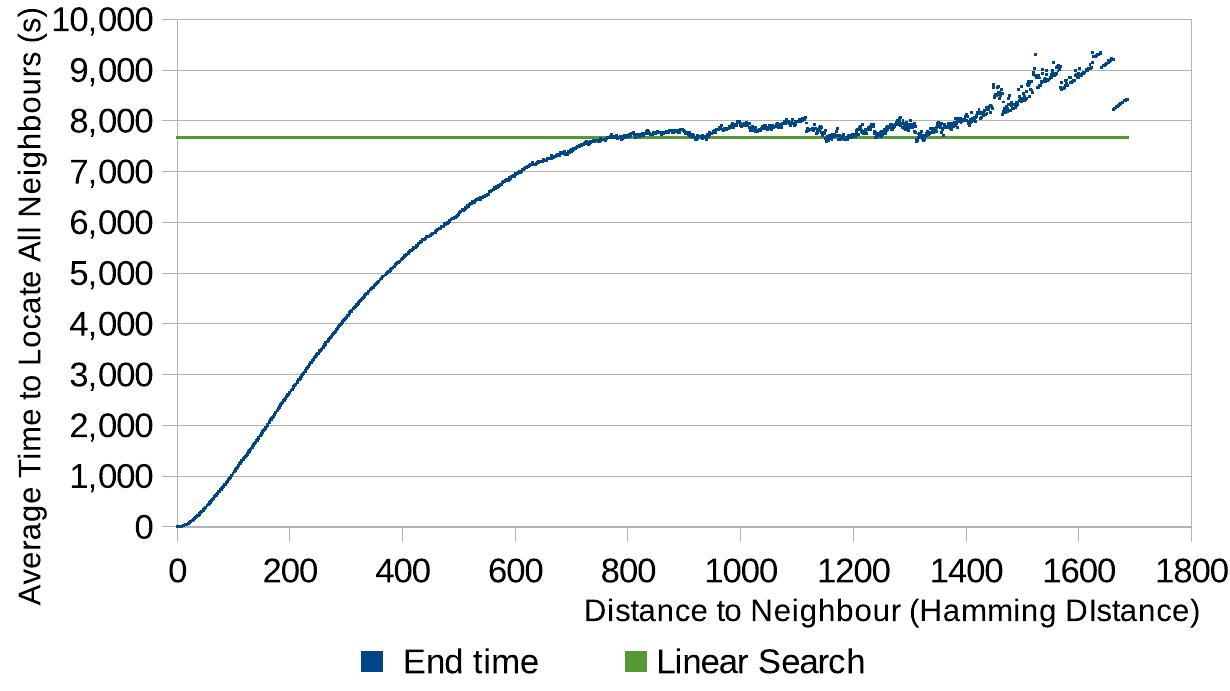}
    \caption{\small Chart showing the average time in seconds required to locate all neighbours up to a given Hamming distance for a Hamming Tree and a linear search.}
    \label{fig:index_query_times}
\end{figure}

While the algorithm can to some extent be parallelised, the testing was done using a single threaded implementation. The time from the start of each query to when all nearest neighbours up to each Hamming distance were located using the Hamming Tree was measured. These timings were averaged across the 1000 queries for every value of Hamming distance. For comparison, the cost of performing a linear search through the set of QUICCI images is also reported; this has a constant time as it has to traverse the entire list of 828.5 million QUICCI images. The results are shown in Figure \ref{fig:index_query_times}. 

The Figure shows that, particularly for small bit distances, the Hamming Tree is very effective at reducing query times. This makes it a good candidate for neighbour discovery when using LSH-derived methods. Query execution times are highly dependent on the presence of close neighbours to a given needle image and the number of search results requested, but generally follow the timing pattern shown.

It is worth noting here that the average number of set bits per QUICCI image in the created dataset was measured to be 610.1. When the Hamming Tree reaches parity with a linear search at a Hamming distance of about 800, the relevance of the search results is not expected to be high. Moreover, the vast majority of the algorithm's execution time is spent on reading and decompressing files stored on disk. This applies to both the sequential and the indexed query implementations. Chappell et al. \cite{chappell2015approximate} performed all searches in memory, which complicates direct comparison of the two implementations.

\subsection{QUICCI Descriptiveness and Clutter Resistance}
\label{sec:quicciPerformance}

The descriptiveness and clutter resistance of the QUICCI descriptor is evaluated using the Clutterbox experiment proposed in \cite{vanrici}, and used to compare the performance of QUICCI against the RICI, SI, 3DSC, and FPFH descriptors. The RICI descriptor was chosen due to its similarity to the QUICCI descriptor, SI and 3DSC for being the most referenced methods known for their clutter resistance \cite{guo2016comprehensive}, and FPFH is an example of a popular descriptor.

The experiment aims to quantify the clutter resistance of the descriptors by measuring the response of a tested descriptor to increasing levels of clutter. The experiment is executed a large number of times by varying objects and their transformations, in order to provide robust results independent of object type. 

The Clutterbox experiment is performed using the following steps:

\begin{enumerate}
    \item Define a cube volume whose edge size is $s$.
    \item From a large object collection, draw $n$ objects at random.
    \item Fit each of the randomly chosen objects in a unit sphere centred around the origin.
    \item From the selected objects, select one at random to be what is referred to as the ``reference object''.
    \item Compute a descriptor for each unique vertex in the reference object, thereby creating the set of reference descriptors $\{RD\}$.
    \item Iterating over the list of chosen objects in a random order, though always starting with the reference object, do the following for each:
    \begin{enumerate}
        \item Place the object at a randomly chosen orientation and position whose bounding sphere fits entirely within the cube volume.
        \item Compute a descriptor for each unique vertex present in the combined mesh present inside the cube volume. The result is a set of cluttered descriptors $\{CD\}$.
        \item For each $d\in\{RD\}$, compute a list of distances for each $c\in\{CD\}$, and sort it. Locate the corresponding cluttered descriptor in this list, and store its rank in the list ($0 \leq rank \leq |\{CD\}|-1$). Note that lower ranks are better, and rank 0 is the front/top of the list.
        \item From the computed list ranks, construct a histogram where bin $i$ stores the number of occurrences where the corresponding cluttered descriptor was found at rank $i$.
    \end{enumerate}
\end{enumerate}

The procedure is repeated for each tested descriptor, where all randomly selected values are kept constant. The result of the experiment is therefore a list of histograms, one for each level of clutter. While performing the experiment, the parameters listed in Table \ref{table:parameters} were used.

\begin{table}[]
\begin{tabular}{|l|l|}
\hline
\textbf{Parameter}          & \textbf{Value}                        \\ \hline
Clutterbox size             & $s = 3$                               \\ \hline
Object counts               & $n = 1$, $n = 5$, $n = 10$            \\ \hline
\begin{tabular}[c]{@{}l@{}}Support radius \\(all descriptors) \end{tabular} & $r = 0.3$ \tablefootnote{Note that all objects are first fit inside a unit sphere.}  \\ \hline
QUICCI resolution           & 63x64 bits \tablefootnote{Corresponds to the equivalent RICI resolution.}   \\ \hline
RICI / SI resolution        & 64x64 pixels                          \\ \hline
SI support angle            & $180^{\circ}$ (disabled) \tablefootnote{We have not found evidence for its claimed benefits.} \\ \hline
\begin{tabular}[c]{@{}l@{}}3DSC minimum \\support radius \end{tabular} & $r_{min} = 0.048$ \tablefootnote{Proportionally equivalent to previous work.} \\ \hline 
3DSC bin dimensions         &  $J = 15$, $K = 11$, $L = 12$ \tablefootnote{Equivalent to previous work \cite{guo2016comprehensive} \cite{frome2004recognizing}.} \\ \hline
FPFH bins per feature       & 11 \tablefootnote{Equivalent to previous work.} \\ \hline
\begin{tabular}[c]{@{}l@{}}Mesh sampling resolution \\ \hfill \end{tabular}    & \begin{tabular}[c]{@{}l@{}}10 point samples \\ per triangle in mesh \tablefootnote{SI, 3DSC, and FPFH require point clouds, necessitating uniform sampling.} \end{tabular} \\ \hline
\end{tabular}
\caption{Parameters that were used during the evaluation of the different methods. }
\label{table:parameters}
\end{table}

The clutterbox experiment was executed 1,500 times on objects from the SHREC2017 dataset \cite{savva2017shrec}, which contains a total of 51,162 triangle meshes. An exception has been made to the FPFH descriptor, which was only executed 500 times due to excessive execution times. For clarity, all curves of this descriptor have been stretched for easier comparison against other descriptors. 

\begin{figure}[H]
    \centering
    \includegraphics[width=8.8cm]{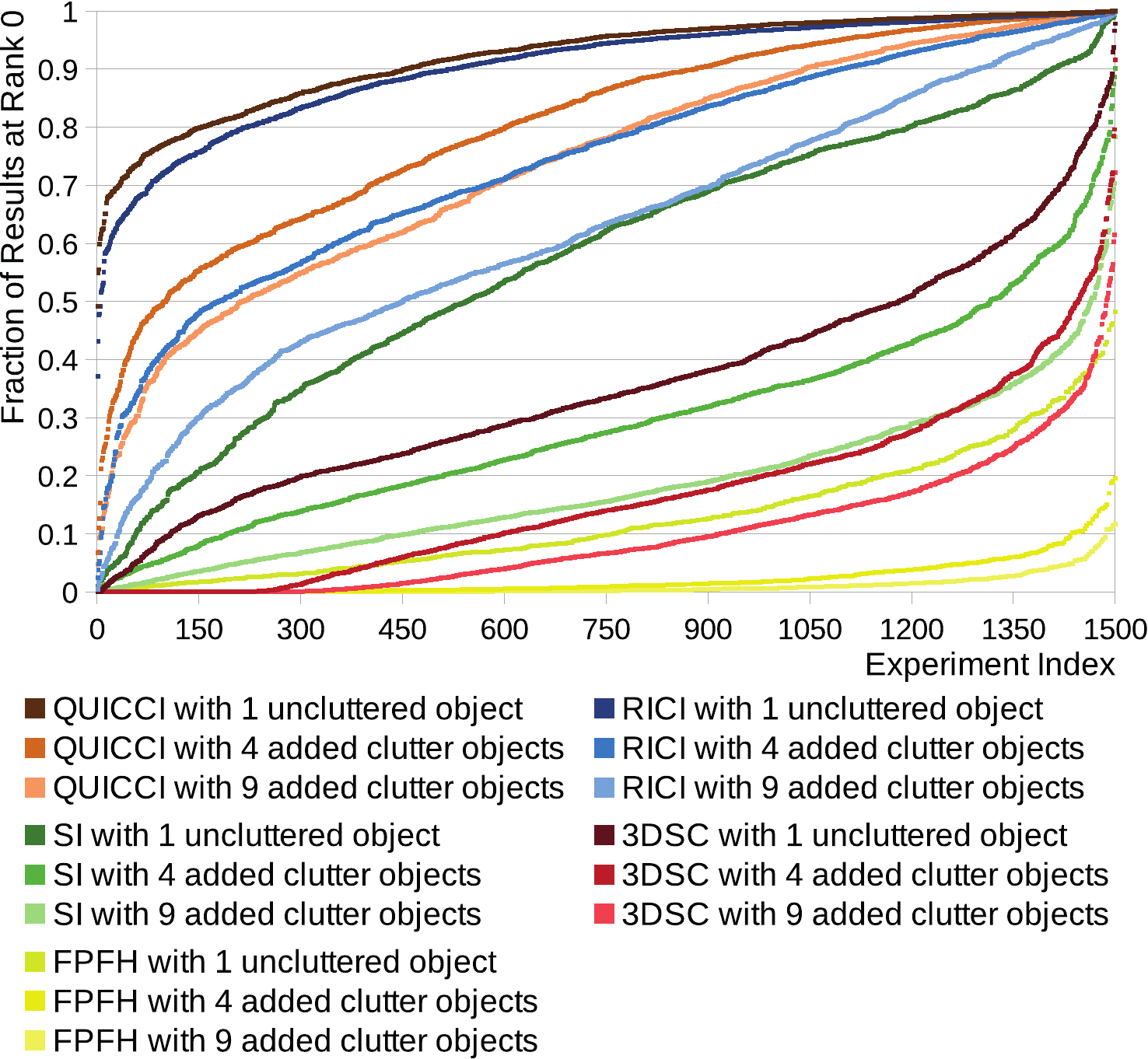}
    \caption{\small Fraction of search results that were correctly ranked at the top of the list of search results for each tested descriptor and added clutter object count. Each sequence has been sorted individually to create monotonically increasing curves.}
    \label{fig:main_performance_chart}
\end{figure}

To visualise the resulting histograms, the fraction of search results correctly being ranked as the best match for each uncluttered reference descriptor (at rank 0) was computed for each descriptor and clutter object count. The produced measurements exhibited a high degree of noise, which did not allow the data to be displayed in a comprehensive manner. Each sequence was therefore sorted individually to produce a monotonically increasing curve, for the purpose of chart readability. The result of this is shown in Figure \ref{fig:main_performance_chart}. The clutterbox experiment was implemented in C++, and the tested descriptors have been implemented on the GPU using CUDA 10.1. 

\begin{figure*}
    \centering
    \begin{subfigure}[t]{18cm}
        \includegraphics[width=18cm]{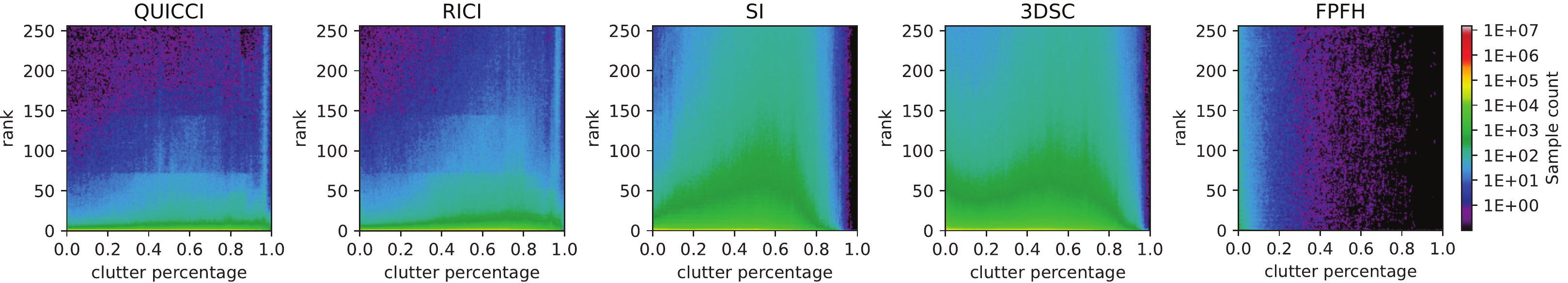}
    \end{subfigure}
    \caption{\small Heatmaps showing the relationship between varying degrees of clutter and each descriptor's matching performance. The horizontal axis represents fraction of area in the support region not part of the matched object, while the vertical axis denotes ranks in the list of search results (where lower is better).}
    \label{fig:heatmaps}
\end{figure*}

The results show that the QUICCI descriptor outperforms previous work in terms of resistance to clutter. However, it is also advantageous to investigate the relationship between the degree of clutter present in the support radius, and the resulting matching performance of each descriptor. To this end, a set of heatmaps was created from the search results of $n = 5$ (4 added clutter objects), showing this relationship. This result set corresponds to a total of 70.0M needle descriptors and associated search results. These can be seen in Figure \ref{fig:heatmaps}. The vertical axis in these heatmaps represents the rank where the correct search result was found (lower rank is better), and the horizontal axis denotes the fraction of clutter (the proportion of surface area in the descriptor's support region not belonging to the object being queried). Higher fractions of clutter generally imply greater difficulty for a given descriptor to correctly identify the correct matching vertex.

From these heatmaps it can be seen that the QUICCI descriptor has similar characteristics to RICI in terms of clutter resistance, albeit with slightly better performance. This reflects the observations from the results of Figure \ref{fig:main_performance_chart}. One possible reason why QUICCI outperforms RICI, is that RICI compares absolute intersection counts, while QUICCI only looks at differences. In the presence of clutter, these absolute values may become noisy, and consequently reduce matching performance.

The FPFH heatmap has a distinct appearance relative to the other methods, which can be attributed to its poor matching performance, particularly in cluttered scenes. The heatmap only counts results that appeared in the top 256 ranks, and shows that even in situations with low fractions of clutter, very few results end up within the top 256 ranks shown in the image. 

\subsection{QUIICI Comparison Rate}

\begin{figure}[H]
    \centering
    \includegraphics[width=8.8cm]{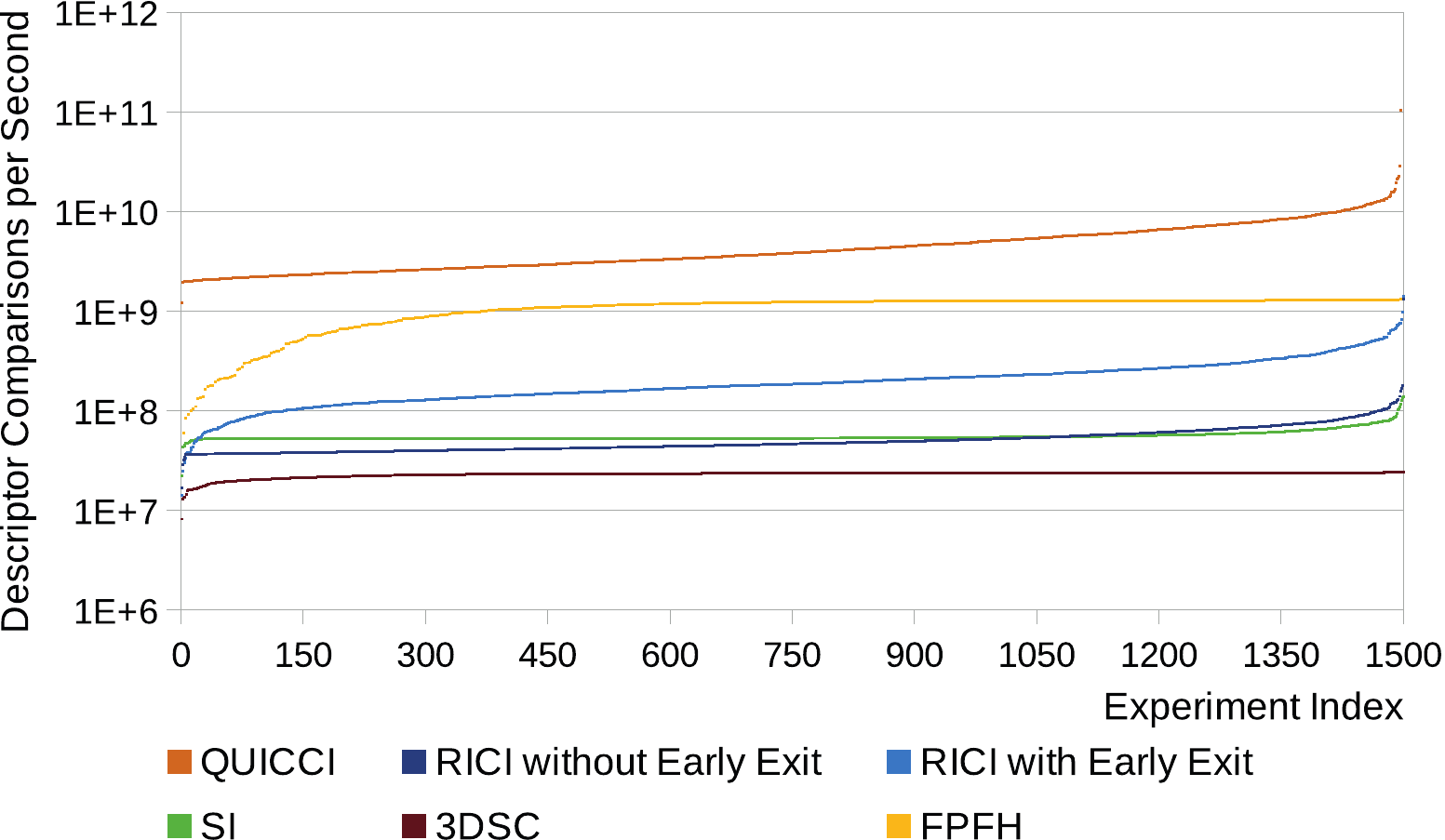}
    \caption{\small Comparison of the number of descriptor pairs each method can compare per second. For readability, each sequence has been sorted individually to create monotonically increasing curves.}
    \label{fig:comparison_rate_chart}
\end{figure}

The number of QUICCI image comparisons performed per second was also measured during the experiments and compared to the other descriptors. In similar fashion to Figure \ref{fig:main_performance_chart}, there was a degree of noise present in the results, and sorting each shown curve individually allowed for the best chart readability. The results are shown in Figure \ref{fig:comparison_rate_chart}. As can be seen, many billions of comparisons can be done per second and, on average, outperforms previous work by over an order of magnitude. This is due to the binary nature of QUICCI.

For the RICI measurements, two variants of the distance function are tested. When an upper distance bound is known, as is often the case in retrieval applications, distance computation can cease early as this value can only grow. Results with early exit disabled serve as a baseline execution time, whereas those with the early exit enabled represent a best case. While this early exit could also be implemented for QUICCI images, it is not expected to improve performance much, if at all, due to the additional instruction overhead. Instruction counts are more relevant for QUICCI than RICI, as many QUICCI bits can be compared with a single bitwise instruction, whereas RICI compares each pixel individually.

\subsection{QUICCI Generation Rate}

\begin{figure}[H]
    \centering
    \includegraphics[width=8.8cm]{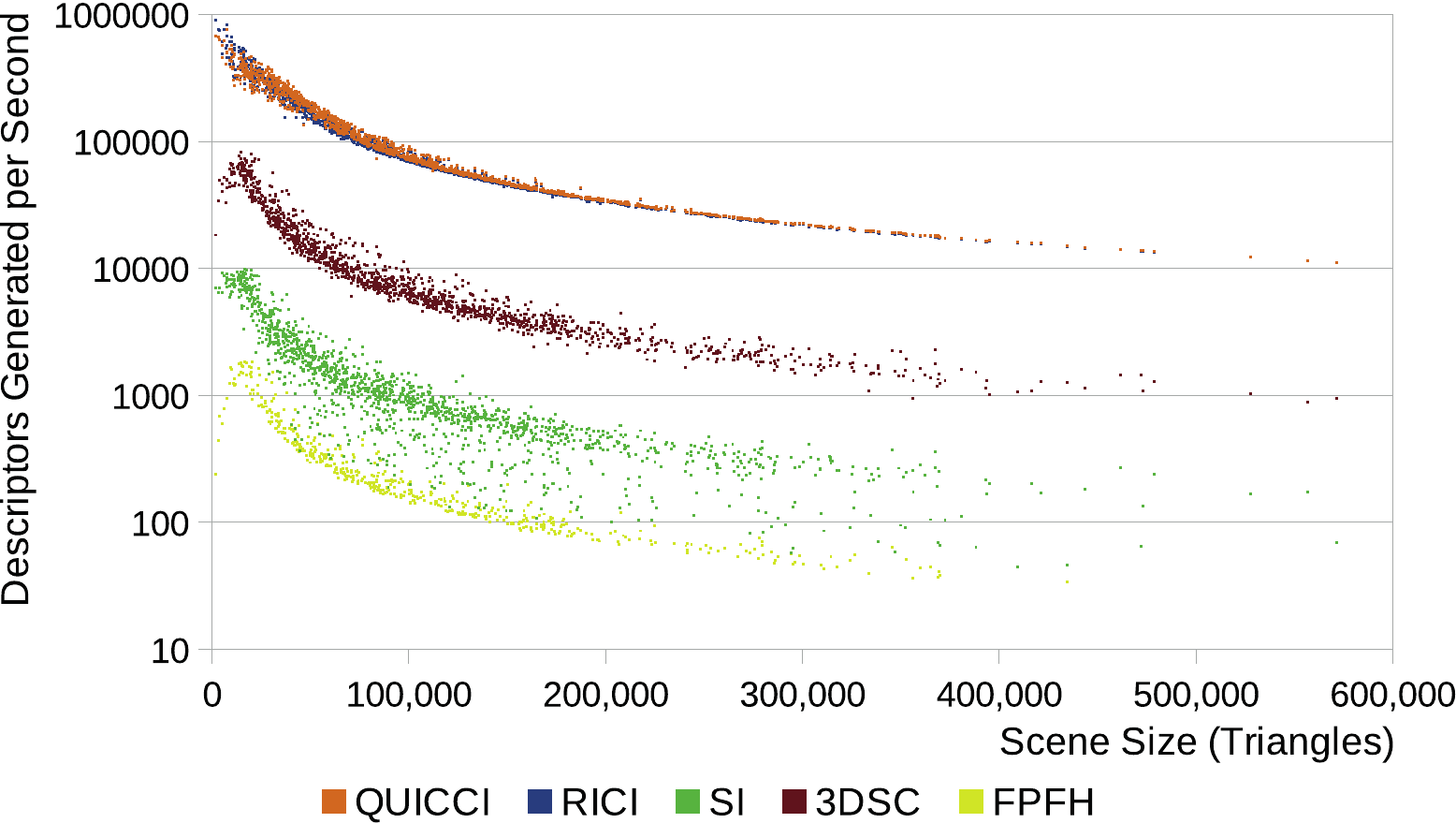}
    \caption{\small A plot showing the relationship between scene size measured in triangles and the rate at which descriptors are computed per second.}
    \label{fig:generation_performance_chart}
\end{figure}

Finally, the rate at which the tested descriptors are computed was measured during the performed experiments. The results are shown in Figure \ref{fig:generation_performance_chart}. The chart shows that QUICCI and RICI descriptors can be generated at similar speeds, which is about an order of magnitude better than the next best descriptor.

\subsection{Weighted Hamming}
\label{sec:whammingEvaluation}

An experiment was constructed in order to quantify our claim that the Weighted Hamming distance function is superior for retrieval purposes of QUICCI images over the clutter resistant distance and Hamming distance functions. The premise of this experiment is to evaluate the distance values returned by each distance function. We compare two different settings: where the surface points being compared have distinctly different support regions and where the support regions are quite similar.

The values returned by the distance functions where object surfaces are distinctly different gives insight into the range of distances that can be expected to be returned by the distance function under ``nominal'' conditions.

\begin{figure}[H]
    \centering
    \includegraphics[width=8.8cm]{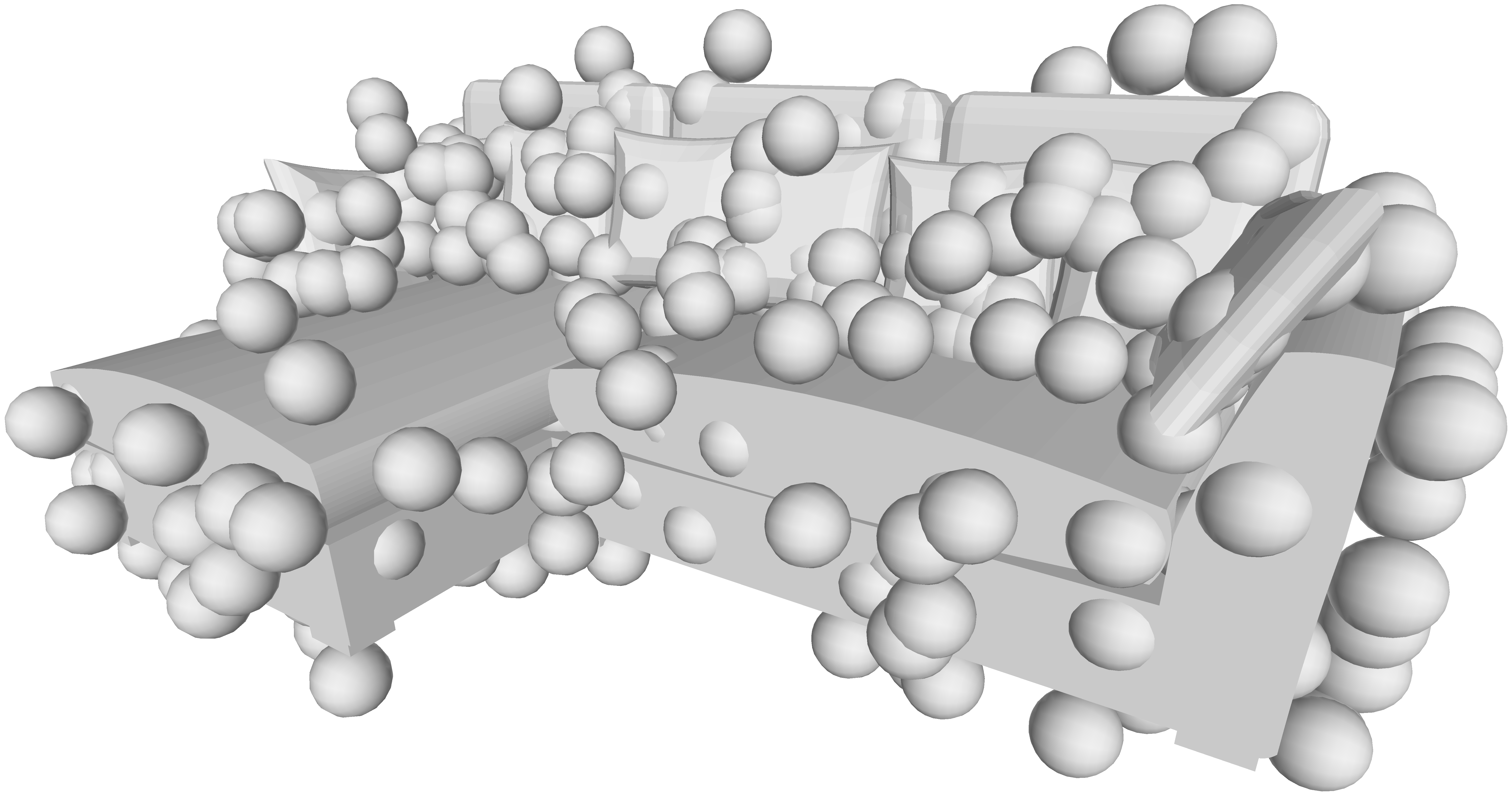}
    \caption{\small A visualisation of an object on which 500 spheres have been placed (the highest number used in the described experiment).}
    \label{fig:spheredBench}
\end{figure}

With that background, one can then investigate what happens to the distance values when point pairs have varying degrees of similarity. In order to obtain quantitative results, it must be possible to generate these varying degrees of similarity automatically \footnote{There exist various ways of generating variants of similar shapes, notably those utilising shape grammars \cite{muller2006procedural} \cite{genevaux2013terrain}. However, the complexity of constructing these shape grammars tends to be high, while the variety of local surfaces produced is often low due to the reuse of a limited set of ``aphabet'' shapes.}. It is possible to simulate variations in geometric similarity through the addition of geometry, whose shape does not necessarily matter. In the devised experiment, spheres are added touching on randomly sampled points on the object's surface. An example of an object with spheres added to its surface in this manner is shown in Figure \ref{fig:spheredBench}.

\begin{figure*}
    \centering
    \begin{subfigure}[t]{18cm}
        \includegraphics[width=18cm]{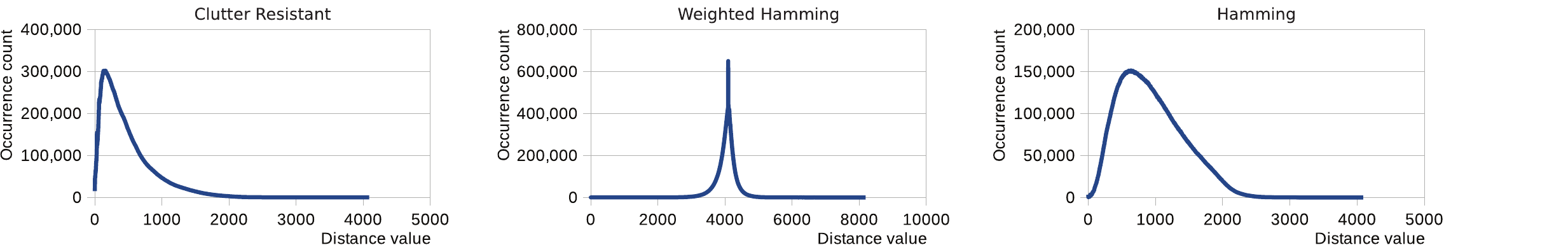}
        \caption{\label{fig:baseline} Distribution of measured distance values under nominal conditions for each tested distance function.}
    \end{subfigure}
    \\
    \begin{subfigure}[t]{18cm}
        \includegraphics[width=18cm]{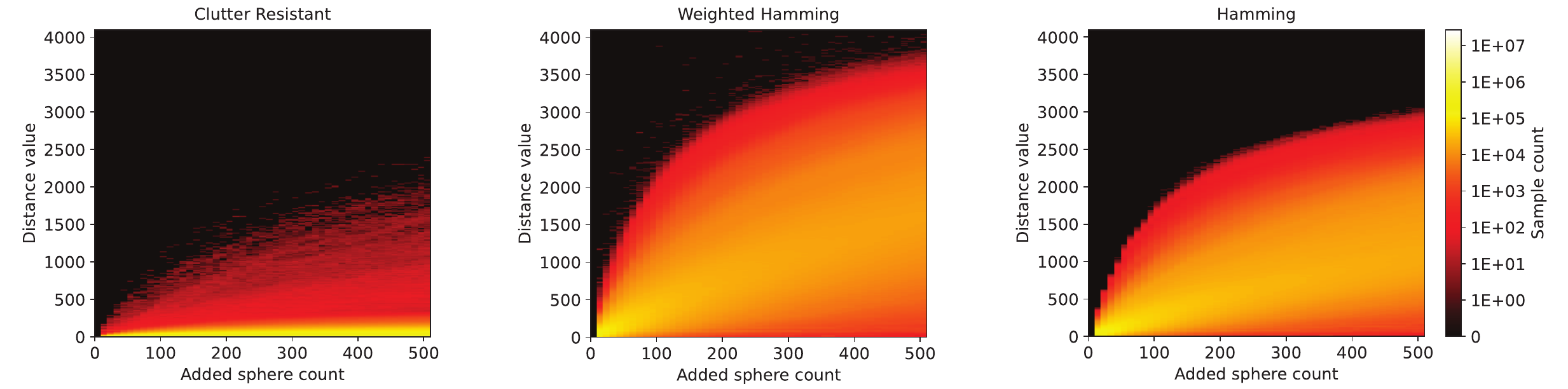}
        \caption{\label{fig:wheatmaps} Visualisation of all 51 distance function response histograms produced when comparing object pairs having varying degrees of surface similarity. Each column represents a single histogram similar to the one shown in Figure \ref{fig:baseline}, corresponding to the distance value distribution when the modified object has a set number of spheres applied to its surface.}
    \end{subfigure} 
    \caption{\small Visualisation of the histograms of distance function responses that were obtained as part of the evaluation of these functions under both nominal and similar surface conditions.}
    \label{fig:weightedHammingHeatmaps}
\end{figure*}

For computing distance values under ``nominal'' conditions, two objects were selected from the (same as previously used) SHREC2017 dataset \cite{savva2017shrec} at random, and scaled to fit within a unit sphere. For each unique vertex in each object, a QUICCI descriptor was computed with the same generation parameters as in Section \ref{sec:quicciPerformance}.
Each pair of QUICCI descriptors corresponding to vertices with the same index\footnote{The number of generated images is bounded by the object with the fewest vertices.} across the 2 objects (which have a random degree of similarity) was used to compute the distance value for each of the 3 distance functions. These values were used to construct the histograms of Figure \ref{fig:baseline}. This process was repeated for 10,000 object pairs, generating 176.2M image pairs.

The next step is to check the stability of the distance functions across the same object vertices as the environment of the vertices is changed. To this end, a random object from the SHREC2017 dataset is selected and fitted within a unit sphere. For each object vertex, a QUICCI descriptor is computed. Next, 10 random points on the object's surface are chosen and normal vectors are computed for these points by interpolation. At each of these points, a sphere of radius 0.05 units is placed such that it touches the selected sample point. This is achieved by displacing the sphere's origin by its radius along the point's normal. After each step of adding 10 spheres (up to a limit of 500 spheres), QUICCI descriptors are computed for the vertices of the original object. Distance values are then computed for each of the 3 distance functions between corresponding vertex QUICCI descriptors of the original and modified objects. After repeating this experiment for 1000 random objects from SHREC2017, histograms of the combined distance values are computed (from a total of 26.79M QUICCI images), see Figure \ref{fig:wheatmaps}.

A good distance function should clearly discriminate between relevant and irrelevant descriptors with respect to a query. For the presented experiment, objects with fewer spheres applied to their surface should be considered closer to their original (unmodified) version by a given distance function. As can be seen from Figure \ref{fig:wheatmaps}, this is indeed the case; all tested distance functions return on average lower scores for objects with fewer spheres.

However, the performance of these distance functions varies when it comes to their ability to discount images not relevant to the needle. For example, when comparing results for Hamming Distance, in Figure \ref{fig:baseline} (right) and \ref{fig:wheatmaps} (right) it can be observed that the histograms (columns of Figure \ref{fig:wheatmaps} (right)) quickly approach the histogram of Figure \ref{fig:baseline} (right) for random vertices.

At a glance, the clutter resistant distance function appears to have the same issue. However, closer inspection of the data shows that the cause of this behaviour is the commonly low number of set bits present in needle images. As the distance function is bounded by the number of set bits present in the image (with the exception of cases where none are set), computed distances have a tendency to be low. However, the vast majority of scores ends up being the highest possible score that the distance function allows for that particular needle image. 

While this behaviour is effective at discerning close matches (as demonstrated in Section \ref{sec:quicciPerformance}), it is less advantageous for retrieval purposes, where more granularity is desirable for the purpose of ranking search results. The Weighted Hamming distance function is the one of the three tested functions which is the most capable of this. Moreover, of the three, it is also the one which shows the clearest separation between distances of matching surfaces relative to distances measured under 'nominal' conditions. A significant amount of variation can be applied before the range of computed scores reaches the same territory as the one shown in the nominal occurrence chart above. Moreover, under these circumstances only a small fraction of results overlaps with this nominal range. For these reasons we conclude that, among the tested distance measures, the Weighted Hamming distance is most suitable for the purposes of retrieval.

\section{Issues with FPFH}
\label{sec:discussion}

Some issues were discovered while testing the well-known FPFH descriptor. The most notable of which pertains to the equation used to construct the FPFH during the second stage of generation, listed in Equation 4 in \cite{rusu2009fast}, reproduced here as Equation \ref{eq:fpfh_construction}.

\begin{equation}
\label{eq:fpfh_construction}
    FPFH(p) = SPF(p) + \frac{1}{k}  \sum_{i = 1}^{k} \frac{1}{ \omega _{k}} \cdot SPF(p_{k}) 
\end{equation}

The Equation computes an FPFH descriptor at point $p$, using a set of $SPF$ histograms that were computed for each point in a previous step, and includes all $k$ neighbours present within the support radius.

Of specific interest here is the distance weighting component $\frac{1}{\omega _{k}}$, which discounts the contribution of each point neighbour's $SPF$ histogram by the distance to the point $p$ for which the $FPFH$ histogram is computed. The issue is that, as this distance is not normalised, the weighting between the left ($SPF(p)$) and right ($\sum_{i = 1}^{k} \frac{1}{ \omega _{k}} \cdot SPF(p_{k}$)) terms of the equation depend on the scale of the object. 

Also worth noting is that the original FPFH paper does not give a distance function to compare descriptors. We have used Pearson correlation in our implementation.

Finally, one detail that we noted in the currently available GPU implementation of Point Cloud Library \cite{Rusu_ICRA2011_PCL} is that the aforementioned weighted distance factor uses the squared distance as a the value of $\omega _{k}$, which deviates from the original paper.

\section{Conclusion}

This paper addressed the problem of querying by local shape. A new binary descriptor, QUICCI, is proposed which is robust to clutter, highly descriptive and quite small in size. To overcome the cost of searching the huge number of such descriptors that result from an object collection, a binary image indexing scheme, the Hamming Tree, was proposed which can significantly accelerate searching, especially for small Hamming distances. The effectiveness of an indexing structure is, however, highly dependent on the distance function used. The Weighted Hamming distance function is also proposed, which can be used to rank QUICCI descriptors in a retrieval setting.

\section{Acknowledgements}

The authors would like to thank the HPC-Lab leader and PI behind the "Tensor-GPU" project, Prof. Anne C. Elster, for access to the Nvidia DGX-2 system used in the experiments performed as part of this paper. Additionally, the authors would like to thank the IDUN cluster at NTNU for the provision of additional computing resources.

\printbibliography

\end{document}